\documentclass[journal]{IEEEtran}

\ifCLASSINFOpdf
\else
   \usepackage[dvips]{graphicx}
\fi
\usepackage{url}

\usepackage{adjustbox}
\usepackage{amsfonts}
\usepackage{amsmath}
\usepackage[bb=dsserif,bbscaled=1.1]{mathalpha}
\usepackage{amssymb}
\usepackage{booktabs}
\usepackage{graphicx}
\usepackage{multirow}

\usepackage[pagebackref,breaklinks,colorlinks]{hyperref}

\usepackage{lipsum}  % for generating dummy text
\usepackage{eso-pic}  % for placing text in the background

\AddToShipoutPictureBG*{%
  \AtPageLowerLeft{%
    \put(\LenToUnit{0.5\paperwidth},\LenToUnit{0.03\paperheight}){%
      \makebox[0pt][c]{\textbf{\footnotesize Approved for public release, NGA-U-2023-02063}}%
    }%
  }%
}

\begin{document}

\title{
Debiased Learning for Remote Sensing Data
}

\author{%
\setlength{\tabcolsep}{5pt}
\begin{tabular}{@{}lcccccr@{}}
Chun-Hsiao Yeh&
Xudong Wang &
Stella X. Yu&
Charles Hill& 
Zackery Steck& 
Scott Kangas&
Aaron Reite\\
\end{tabular}

\thanks{Chun-Hsiao Yeh, Xudong Wang, Stella X. Yu are with the University of California, Berkeley, CA 94720 USA (\{daniel\_yeh, xdwang\}@berkeley.edu). Yu is also with the University of Michigan, Ann Arbor, MI 48109, USA (stellayu@umich.edu).
Charles Hill, Zackery Steck, Scott Kangas are with Etegent Technologies, Ltd. (\{charles.hill,zackery.steck,scott.kangas\}@etegent.com).
Aaron Reite is with the National Geospatial-Intelligence Agency (aaron.a.reite@nga.mil).
}
}

\markboth{}%Journal of \LaTeX\ Class Files}
{Shell \MakeLowercase{\textit{et al.}}: Bare Demo of IEEEtran.cls for IEEE Journals}
\maketitle

\begin{abstract}
Deep learning has had remarkable success at analyzing handheld imagery such as consumer photos due to the availability of large-scale human annotations (e.g., ImageNet). However, remote sensing data lacks such extensive annotation and thus potential for supervised learning. To address this, we propose a highly effective semi-supervised approach tailored specifically to remote sensing data. Our approach encompasses two key contributions. First, we adapt the FixMatch framework to remote sensing data by designing robust strong and weak augmentations suitable for this domain. Second, we develop an effective semi-supervised learning method by removing bias in imbalanced training data resulting from both actual labels and pseudo-labels predicted by the model. Our simple semi-supervised framework was validated by extensive experimentation. Using 30\% of labeled annotations, it delivers a 7.1\% accuracy gain over the supervised learning baseline and a 2.1\% gain over the supervised state-of-the-art CDS method on the remote sensing xView dataset.
\end{abstract}


\IEEEpeerreviewmaketitle

\section{Introduction and Related Works}
\IEEEPARstart{D}{eep} learning is remarkably successful at analyzing natural images, primarily due to the extensive human annotations available in datasets such as ImageNet~\cite{krizhevsky2017imagenet}. However, the scenario changes significantly when using remote sensing data~\cite{lam2018xview,sun2022fair1m,xia2018dota}. While there is an abundance of remote sensing data available, there are limited annotations due to the challenges in annotating such data: uncommon viewing angles and ambiguity in remote sensing data make it difficult for humans to annotate accurately.

\begin{figure}[t]
    \centering
    \includegraphics[width=0.95\linewidth]{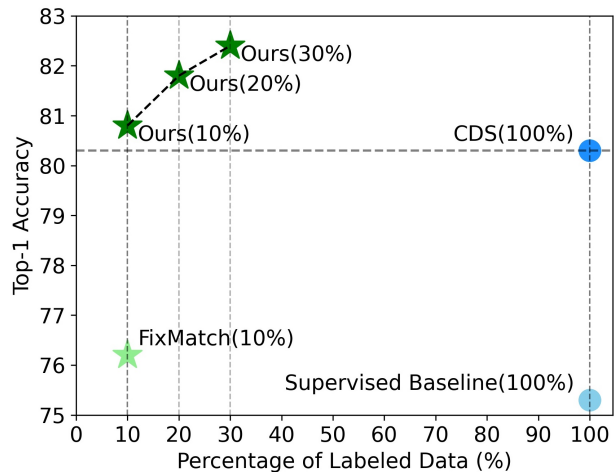}
    %\vspace*{-3mm}
    \caption{Our semi-supervised learning approach outperforms FixMatch~\cite{sohn2020fixmatch}, supervised baseline~\cite{singhal2022multi}, and CDS~\cite{singhal2022multi} on remote sensing data~\cite{lam2018xview}.
    }
    \label{fig:teaser}
    %\vspace{-5mm}
\end{figure}

Previous works, such as transfer learning approaches, have had significant success in handling specific data domains by starting with a model pretrained on ImageNet~\cite{krizhevsky2017imagenet} and fine-tuning it with a few samples of the target domain. These approaches have been highly effective for medical imaging~\cite{rajpurkar2017chexnet} and self-driving~\cite{kim2017end} tasks. However, unlike these image domains, remote sensing data is characterized by a lack of high-resolution details and class-imbalanced properties, making it challenging to transfer from an ImageNet pretrained model directly~\cite{thirumaladevi2023remote,fuller2022transfer,pires2019convolutional,chen2020big}.

Additionally, several proposed self-supervised learning methods~\cite{wu2018unsupervised, he2020momentum, chen2020simple, grill2020bootstrap} can learn representations without human-annotated data. However, learning self-supervised models without pretraining is expensive and time-consuming. Moreover, the ability to discriminate positive and negative pairs may be worse on remote sensing data than on natural images due to the presence of complex and ambiguous patterns within the data.

In remote sensing scenarios with limited annotated data and abundant unlabeled data, semi-supervised learning techniques~\cite{berthelot2019mixmatch, berthelot2019remixmatch, lee2013pseudo, li2021comatch} hold significant promise. FixMatch~\cite{sohn2020fixmatch} is a popular approach that combines weak and strong augmentations with pseudo-labeling to enhance the model's performance with limited labeled data. However, to adapt these techniques to remote sensing domains, we focus on the design of robust weak and strong augmentations specifically tailored to this context. We investigate the benefits of incorporating rotation, scaling, and horizontal flipping, with the aim of unlocking the full potential of semi-supervised learning for remote sensing data, enabling the use of more effective and context-aware analysis techniques.

\begin{figure}[t]
    \centering
    \includegraphics[width=0.75\linewidth]{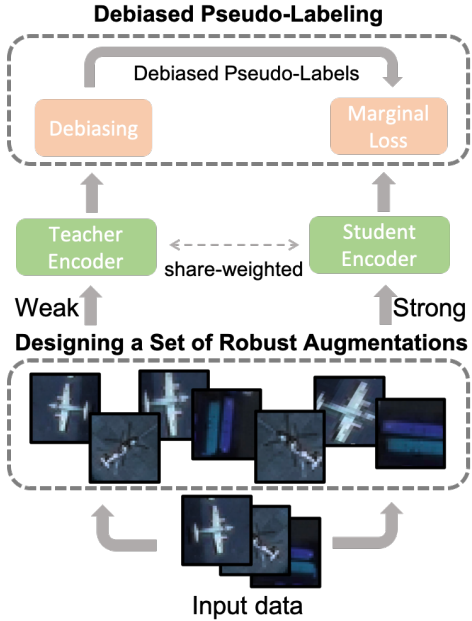}
    %\vspace*{-3mm}
    \caption{Our semi-supervised learning framework. The framework consists of two key components. The first component focuses on the design of robust strong and weak augmentations specifically tailored for remote sensing data. 
The second component leverages a state-of-the-art debiased learning approach~\cite{wang2022debiased} to effectively address the bias present in pseudo-labeling. 
    }
    \label{fig:model}
    %\vspace{-3mm}
\end{figure}

Previous studies have identified two sources of imbalance in real-world data~\cite{van2018inaturalist, gupta2019lvis} which pose significant challenges to training deep neural networks. First is the class-imbalance problem which various works~\cite{cui2019class, park2021influence, cao2019learning, wang2020long} have attempted to address. Furthermore, machine-generated pseudo-labels often suffer from inherent imbalances~\cite{wang2022debiased}. These imbalances pose a challenge for learning models as they introduce a bias towards false majorities within the pseudo-labels. Similar issues have been observed in widely used datasets like CIFAR and ImageNet~\cite{wang2022debiased}. To address this pseudo-label bias problem specifically in the context of remote sensing data, we incorporate the DebiasPL~\cite{wang2022debiased} method, which aims to alleviate the bias associated with pseudo-labeling. We demonstrate the
effectiveness of our approach in Fig.~\ref{fig:teaser}, and illustrate our proposed framework in Fig.~\ref{fig:model}. Our key contributions in this work are summarized as follows:
\begin{enumerate}
%\vspace{-0.5em}
\item We adapt the framework of semi-supervised learning approaches, such as FixMatch, to remote sensing data by designing a set of robust strong and weak augmentations specifically tailored to this domain.

\item We leverage the recently proposed DebiasPL method to mitigate the imbalance in pseudo-labeled data.

\item We conduct extensive experiments that demonstrate the effectiveness of our approach. As shown in Fig.~\ref{fig:teaser}, using 30\% of labeled annotations, our simple semi-supervised framework increases accuracy by 7.1\% compared to the supervised learning baseline and outperforms the supervised, state-of-the-art CDS technique by 2.1\% on the remote sensing xView dataset.

\end{enumerate}

\begin{figure}[t]
    \centering
    \includegraphics[width=1\linewidth]{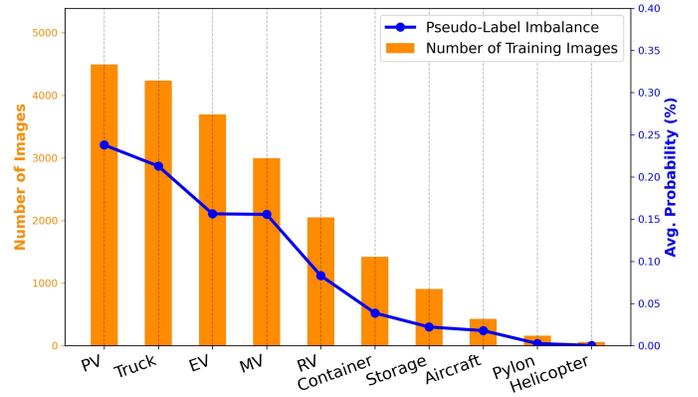}
    %\vspace*{-4mm}
    \caption{
    Remote sensing data, such as xView~\cite{lam2018xview}, has two sources of imbalance. One is the training label imbalance as provided by humans; the other is the pseudo-label imbalance generated by the semi-supervised learning framework during learning on remote sensing data. 
    }
    \label{fig:two_source_imbalance}
    %\vspace{-3mm}
\end{figure}

\section{Debiased Semi-Supervised Learning}
\label{sec:meths}
Our work reviews recent advances in addressing data imbalance in remote sensing images, focusing on FixMatch~\cite{sohn2020fixmatch}, a semi-supervised approach using pseudo-labeling. 

\subsection{FixMatch}

Recent semi-supervised learning approach, FixMatch stands out for its superior performance, driven by a dual-loss mechanism $\mathcal{L} = \mathcal{L}_s + \lambda \mathcal{L}_u$. The first loss $\mathcal{L}_s=\frac{1}{B^l} \sum_{i=1}^{B^l} \mathcal{H}\left(y_i^l, f\left(\mathcal{W}\left(\boldsymbol{x}_i^l\right) ; \boldsymbol{\theta}\right)\right)$, operates on a small amount of labeled data $\mathcal{X}=\left\{\left(\boldsymbol{x}_i^l, y_i^l\right) ; i \in\left(1, \ldots, B^l\right)\right\}$, using weak augmentation $\mathcal{W}(\cdot)$ to preserve original data characteristics. This yields predictions closely aligned with the unaltered instances. 

In the realm of a large amount of unlabeled data $\mathcal{U}=\left\{\boldsymbol{x}_j^u ; j \in\left(1, \ldots, B^u\right)\right\}$, the second loss $\mathcal{L}_u=\frac{1}{B^u} \sum_{j=1}^{B^u} \mathbb{1}\left(\hat{y}_j^u \geq \tau\right) \mathcal{H}\left(\hat{y}_j^u, f\left(\mathcal{S}\left(\boldsymbol{x}_j^u\right) ; \boldsymbol{\theta}\right)\right)$, employs strong augmentations $\mathcal{S}(\cdot)$, granting the model diverse perspectives for enhanced robustness. These augmentations lead to pseudo-labels $\hat{y}_j^u = \max(f(\mathcal{W}(\boldsymbol{x}_j^u); \boldsymbol{\theta}))$, governed by a confidence threshold $\tau$. FixMatch's amalgamation of these losses optimally leverages both labeled and unlabeled data, surpassing previous methods in semi-supervised learning.

\subsection{Sources of Imbalance in Remote Sensing Data}
One challenge with using semi-supervised learning approaches based on pseudo-labeling is the imbalanced nature of both the training data and the pseudo-labels. 

In order to gain a deeper understanding of this issue, in Fig.~\ref{fig:two_source_imbalance}, we analyze the distribution of the training data and pseudo-labels generated by FixMatch~\cite{sohn2020fixmatch} on xView~\cite{lam2018xview}, a remote sensing dataset. Our investigation reveals that these imbalances in the data source can introduce notable biases during the learning process, ultimately hindering the model's performance. Thus, our work aims to:

\begin{enumerate}
\item apply the debiased learning method~\cite{wang2022debiased} to alleviate the imbalanced \textbf{pseudo-labels}, and

\item incorporate logit adjustment~\cite{menon2020long} to mitigate the imbalanced \textbf{training labels} in the remote sensing data.

\end{enumerate}

\subsection{Debiased Learning (Pseudo-Labels)}
Our approach utilizes the DebiasPL method~\cite{wang2022debiased}, which extends the FixMatch framework with an adaptive debiasing module and a marginal loss to mitigate biases in pseudo-labels.

\noindent \textbf{Adaptive debiasing module.} 
Pseudo-labeling often introduces uncertainty due to ambiguous pseudo-label assignments. This can harm model performance, particularly with imbalanced data distribution or biased pseudo-labeling. DebiasPL draws inspiration from Causal Inference~\cite{greenland1999confounding, pearl2009causal, rubin2019essential} and counterfactual reasoning to minimize pseudo-labeling bias impact. It updates the logits of weakly-augmented unlabeled instances as follows:

\begin{equation}
\tilde{f}_i = f\left(\mathcal{W}\left(x_i\right)\right) - \lambda \log \hat{p}
\end{equation}
where $\hat{p}$ is the momentum-updated average of pseudo-label probabilities, and $\lambda$ controls the strength of debiasing.

\noindent \textbf{Adaptive marginal loss.} To address inter-class confusion in pseudo-labels, DebiasPL creates a larger margin between highly biased classes to enforce clearer class boundaries, especially for highly biased classes. It is expressed as:
\begin{equation}
\mathcal{L}_{\mathrm{margin}} = -\log \frac{e^{\left(z_{\hat{y_i}}-\Delta_{\hat{y_i}}\right)}}{e^{\left(z_{\hat{y_i}}-\Delta_{\hat{y_i}}\right)}+\sum_{k \neq \hat{y_i}}^C e^{\left(z_k-\Delta_k\right)}}
\end{equation}
where $\Delta_j=\lambda \log \left(\frac{1}{\hat{p}_j}\right) \text{for } j \in\{1, \ldots, C\}, z=f\left(\beta\left(x_i\right)\right)$.

By incorporating the adaptive debiasing module and the marginal loss into the FixMatch framework, DebiasPL provides a robust approach to alleviate the biases in pseudo-labels and improve the performance of semi-supervised learning on remote sensing data.

\subsection{Logit Adjustment (Training Labels)}
Logit adjustment~\cite{menon2020long} is utilized to address imbalanced training data, which can cause biased performance for majority and minority classes, impacting the classifier's ability to distinguish between them~\cite{singhal2022multi}. During evaluation, this technique adjusts the logits of each class based on their frequencies in the training data. By doing so, the classifier achieves a more balanced performance across all classes in remote sensing data.

\subsection{Our New Findings on Remote Sensing Data}
Remote sensing data possesses unique properties that set it apart from natural images, such as non-uniform ground sampling distance (GSD)~\cite{lam2018xview}. Despite the impressive success of debiased learning with natural images~\cite{wang2022debiased}, it remains uncertain whether these techniques can be effectively applied to remote sensing data. 

Furthermore, the best data augmentation strategies for remote sensing data and the necessary modifications for optimal performance have not been fully explored in debiased learning~\cite{wang2022debiased} or related works~\cite{sohn2020fixmatch,singhal2022multi}. Our research addresses these gaps by investigating the effectiveness of debiased learning techniques on remote sensing data and identifying appropriate modifications and practices to achieve superior performance.

\begin{figure}[ht]
    \centering
    \begin{tabular}{@{}cccc@{}}
        \adjustbox{width=0.08\textwidth}{\includegraphics{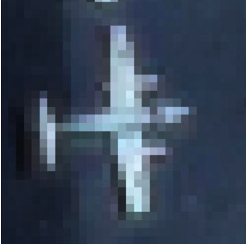}} &
        \adjustbox{width=0.08\textwidth}{\includegraphics{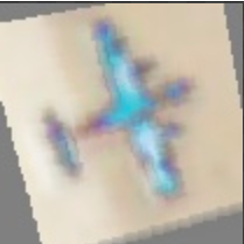}} &
        \adjustbox{width=0.08\textwidth}{\includegraphics{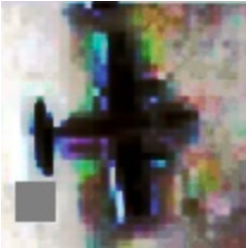}} &
        \adjustbox{width=0.08\textwidth}{\includegraphics{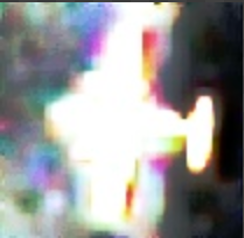}} \\
        \adjustbox{width=0.08\textwidth}{\includegraphics{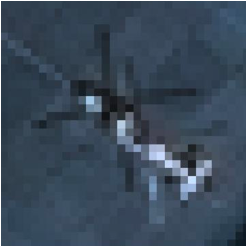}} &
        \adjustbox{width=0.08\textwidth}{\includegraphics{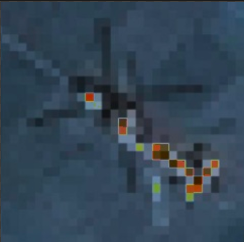}} &
        \adjustbox{width=0.08\textwidth}{\includegraphics{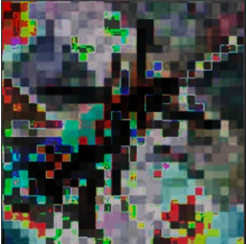}} &
        \adjustbox{width=0.08\textwidth}{\includegraphics{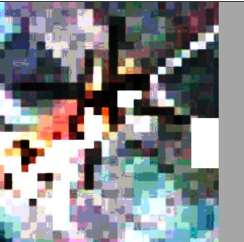}} \\
        \adjustbox{width=0.08\textwidth}{\includegraphics{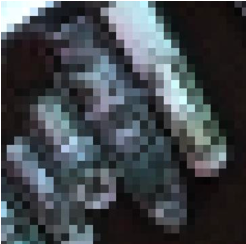}} &
        \adjustbox{width=0.08\textwidth}{\includegraphics{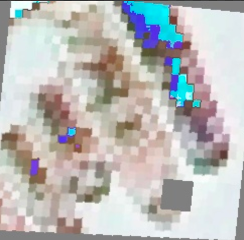}} &
        \adjustbox{width=0.08\textwidth}{\includegraphics{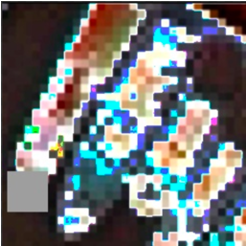}} &
        \adjustbox{width=0.08\textwidth}{\includegraphics{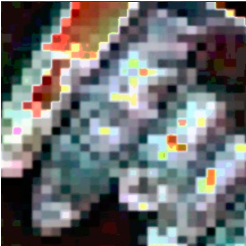}} \\
        \footnotesize Original & \footnotesize RandAugment & \footnotesize RandAugment & \footnotesize RandAugment \\
        &\footnotesize($m$ = 6)&\footnotesize($m$ = 8)&\footnotesize($m$ = 10)
    \end{tabular}
    \caption{\textit{RandAugment}~\cite{cubuk2020randaugment} appears to be too complicated for remote sensing data~\cite{lam2018xview} with artifacts. A series of random transformations are applied to orginal images (Column 1). By increasing the transformation magnitude in terms of parameter $m$, which controls transformation intensity, we can observe more pronounced and visually diverse augmentations, such as artifacts and distortions at the border of image.}
    \label{fig:randaugment}
\end{figure}

\begin{figure}[ht]
    \centering
    \begin{tabular}{@{}ccccc@{}}
        \adjustbox{width=0.073\textwidth}{\includegraphics{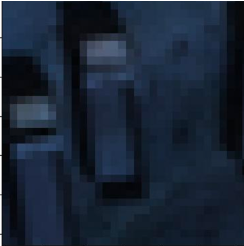}} &
        \adjustbox{width=0.073\textwidth}{\includegraphics{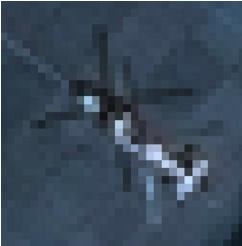}} &
        \adjustbox{width=0.073\textwidth}{\includegraphics{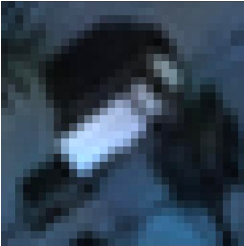}} &
        \adjustbox{width=0.073\textwidth}{\includegraphics{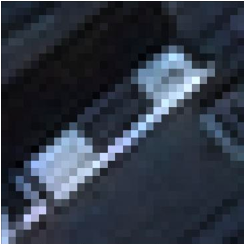}} &
        \adjustbox{width=0.073\textwidth}{\includegraphics{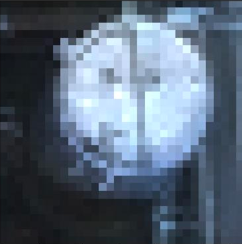}} \\
        \footnotesize EV & \footnotesize Helicopter & \footnotesize PV & \footnotesize RV & \footnotesize Storage \\ 
        \adjustbox{width=0.073\textwidth}{\includegraphics{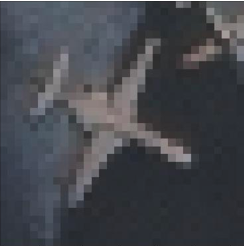}} &
        \adjustbox{width=0.073\textwidth}{\includegraphics{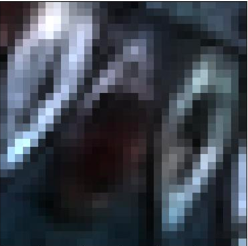}} &
        \adjustbox{width=0.073\textwidth}{\includegraphics{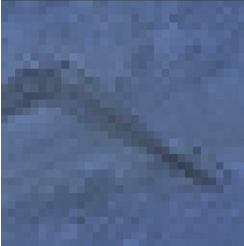}} &
        \adjustbox{width=0.073\textwidth}{\includegraphics{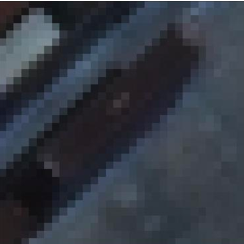}} &
        \adjustbox{width=0.073\textwidth}{\includegraphics{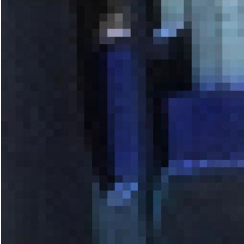}} \\
        \footnotesize Aircraft & \footnotesize MV & \footnotesize Pylon & \footnotesize Container & \footnotesize Truck 
    \end{tabular}
    \caption{{xView data overview.}
    The data contains ten individual classes and has the properties of a highly imbalanced distribution (also shown in Fig.~\ref{fig:two_source_imbalance}).}
    \label{fig:dataset}
\end{figure}

\noindent \textbf{Data augmentations for remote sensing datasets.} In semi-supervised learning (SSL) settings, the debiased learning method~\cite{wang2022debiased} has been shown to be effective using \textit{random resize cropping} and \textit{horizontal flipping} for weak augmentation and \textit{RandAugment}~\cite{cubuk2020randaugment} for strong augmentation on natural images. However, these augmentation techniques do not generalize well to remote sensing data due to the distinct properties of the latter, such as a lack of details in the imagery. For example, in Fig.~\ref{fig:randaugment} we illustrate applications of \textit{RandAugment}~\cite{cubuk2020randaugment} at varying degrees of intensity to remote sensing data, where the method appears to be too severe for the domain.
To address this issue, we empirically select appropriate augmentations that can improve the SSL performance on remote sensing data. Our approach includes \textit{random resize cropping} and  \textit{horizontal flipping} for weak augmentation, and  \textit{horizontal flipping, rotation,} and \textit{scaling} for strong augmentation.

We design the augmentations pipeline for use with remote sensing data and provide rationale for their selection. \textit{Rotation} and \textit{scaling} were both used to simulate variations in imaging angles and ground sampling distances, respectively, which are important factors in generalizing to remote sensing data. Additionally, horizontal flipping was included to simulate mirror-reflected scenes. \textit{RandAugment} was avoided due to the potential to overfit on the remote sensing data~\cite{xia2018dota}. 

The effectiveness of each augmentation method varies. In Table~\ref{tab:augmentation}, we quantify the performance change from each augmentation strategy described in this section. One notable consequence of this study is that the augmentations used in previous SSL works, such as debiased learning, are not always beneficial to remote sensing images.

\section{Experiments}
\label{sec:exp}

We evaluate the effectiveness of our semi-supervised learning approach on the xView remote sensing dataset (Fig.~\ref{fig:dataset}) by comparing the instance-wise accuracy and class-wise accuracy of our model and baselines. Additionally, we conduct ablation studies on data augmentation, data preprocessing, and learning recipe to optimize our method for remote sensing data. All experiments are performed on a single machine using four Nvidia RTX 2080 Ti GPUs.

\subsection{Experimental Setup}
\noindent We evaluate our semi-supervised learning approach on the xView~\cite{lam2018xview} dataset, which contains large-scale multi-spectral images with 8-band channels obtained from satellite data.

We preprocess the data by selecting a subset of 10 categories from the original 60 classes, including \textit{StorageTank, Helicopter, Pylon, Maritime Vessel (MV), ShippingContainer, Fixed-Wing Aircraft (FWAircraft), Passenger Vehicle (PV), Truck, Railway Vehicle (RV), and Engineering Vehicle (EV)}. This dataset (Fig.~\ref{fig:dataset}) consists of 20,431 training samples, 2,270 validation samples, and 63,279 test samples. It is important to note that we only use the R, G, and B bands from the original eight bands to form RGB images.

\noindent \textbf{Baseline methods.} 
We compare our approach against FixMatch~\cite{sohn2020fixmatch} as it is a straightforward and effective semi-supervised learning baseline. Additionally, we also include a ResNet-18 supervised learning baseline and results from CDS~\cite{singhal2022multi}, a complex-valued model recently proposed for remote sensing tasks, for comparison.

\noindent \textbf{Training and evaluation.} 
In the semi-supervised learning setup, we use a ResNet-50~\cite{he2016deep} as the backbone and employ the same hyperparameters as FixMatch. We initialize the model using the pretrained ImageNet model trained with MoCo v2 and EMAN~\cite{cai2021exponential} for 800 epochs and then fine-tune it on the xView dataset for another 150 epochs. For the supervised learning baselines and CDS~\cite{singhal2022multi}, we use a ResNet-18 as the backbone and do not apply data augmentation during training. We measure the top-1 accuracy using both instance-wise and class-wise accuracy metrics to assess the model's performance on imbalanced classes.

\subsection{Instance-wise Comparisons}
Table~\ref{tab:compare_baseline} shows a comparison of our approach against semi-supervised and supervised learning baselines. In our initial setup, we had our framework learn on 10\% labeled data with a ResNet-50 backbone pretrained on ImageNet. Our approach achieved 80.8\% accuracy, outperforming the semi-supervised learning baseline, FixMatch, by 4.6\% and the supervised learning baseline by 5.5\%. 

\begin{figure}[t]
    \centering
    \includegraphics[width=0.75\linewidth]{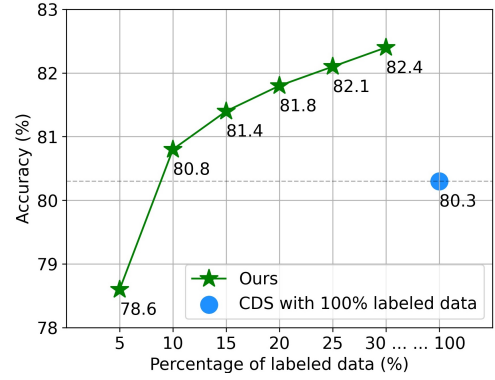}
    %\vspace*{-3mm}
    \caption{{ Improve the performance of our semi-supervised learning framework by increasing the amount of labeled data (5\% to 30\%).}
    }
    \label{fig:label_perc}
    %\vspace{-3mm}
\end{figure}

Moreover, our method outperformed the best setting of the state-of-the-art complex-valued model for remote sensing, CDS~\cite{singhal2022multi}, which was trained with 100\% labeled data, by 0.5\%. In Fig.~\ref{fig:label_perc}, we demonstrate that the performance of our semi-supervised learning framework can be further improved by increasing the amount of labeled data, surpassing the fully-supervised CDS approach.

\begin{table}[t!]
 \centering
  \caption{Top-1 accuracy comparison ($\%$) on remote sensing xView.}
 \resizebox{\columnwidth}{!}{
 \begin{tabular}{lcccccc}
     \toprule
     Method & Pretrained? & Labels (\%) & Backbone & Augmented? & Top-1 (\%) \\
     \midrule
     \midrule
     Supervised~\cite{singhal2022multi} & $\times$ & 100 & RN-18 & $\times$ & 75.3 \\
     CDS~\cite{singhal2022multi} & $\times$ & 100 & RN-18 & $\times$ & 80.3 \\
     \midrule
     FixMatch~\cite{sohn2020fixmatch} & \checkmark & 10 & RN-50 & \checkmark & 76.2 \\
     Ours & \checkmark & 10 & RN-50 & \checkmark & \textbf{80.8} \\
     Ours & \checkmark & 30 & RN-50 & \checkmark & \textbf{82.4} \\
     \bottomrule     
 \end{tabular}
 
 }

 \label{tab:compare_baseline}
\end{table}

\subsection{Ablation Studies}
We conducted an ablation study to assess the effectiveness of selected augmentation strategies on the xView~\cite{lam2018xview} remote sensing dataset. Given the complexity of \textit{RandAugment}~\cite{cubuk2020randaugment} in DebiasedPL, we carefully chose and adapted augmentations suitable for this domain to enhance the model's performance.

Table~\ref{tab:augmentation} summarizes the performance gain achieved with the selected augmentations on xView. Our findings indicate that the adapted augmentations effectively improve the model's performance, resulting in a top-1 accuracy of 80.8\%, a 5.2\% gain compared to the original DebiasPL augmentations. These results highlight the significance of choosing and customizing appropriate augmentations for remote sensing datasets, leading to significant accuracy improvements in semi-supervised learning models on remote sensing data.

\begin{table}[htbp]
    \centering
    \caption{Performance gain from selected augmentations}
    \label{tab:augmentation}
    \resizebox{\linewidth}{!}{%
    \begin{tabular}{@{}lccccccc@{}}
        \toprule
        \textbf{Weak augmentation} \\
        \midrule
        \textit{None} & \checkmark & $\times$ & $\times$ & \checkmark & $\times$ & $\times$ \\
        \textit{Random Resize Cropping} & $\times$ & \checkmark & \checkmark & $\times$ & \checkmark & \checkmark \\
        \textit{Horizontal flipping} & $\times$ & \checkmark & \checkmark & $\times$ & \checkmark & \checkmark \\
        \midrule
        \textbf{Strong augmentation} \\
        \midrule
        \textit{ResizeCropping + Horizontal flipping} & \checkmark & \checkmark & \checkmark & \checkmark & \checkmark & \checkmark \\
        \textit{Rotation($\pm$ 10 degrees)} & $\times$ & $\times$ & $\times$ & \checkmark & \checkmark & \checkmark \\
        \textit{Scaling(0.8, 1.2)} & $\times$ & $\times$ & $\times$ & \checkmark & $\times$ & \checkmark \\
        \textit{RandAugment(m=10)} & \checkmark & \checkmark & $\times$ & $\times$ & $\times$ & $\times$ \\
        \textit{RandAugment(m=5)} & $\times$ & $\times$ & \checkmark & $\times$ & $\times$ & $\times$ \\
        \midrule
        \textbf{Top-1 accuracy (\%)} & 69.8 & 75.6 & 76.3 & 79.4 & 79.7 & \textbf{80.8} \\
        \bottomrule
    \end{tabular}}

\end{table}

\vspace{-3mm}

\section{Conclusion}
We introduced a tailored semi-supervised approach for remote sensing data, tackling training label and pseudo-label imbalances. Our contributions involve adapting FixMatch with customized augmentations and using debiased learning to counter pseudo-label bias. Our results underscore the potential of this approach, achieving remarkable performance gains with just 30\% annotations. 

\section*{Acknowledgment}
This research was supported, in part, by the National Geospatial Intelligence Agency and Etegent Technologies Ltd. We thank Anna Kay for proofreading the article.

{\small
\bibliographystyle{IEEEtran}
\bibliography{egbib}
}

\end{document}